\documentclass[conference]{IEEEtran}

\usepackage{amsmath,amssymb}
\usepackage{graphicx}
\usepackage{booktabs}
\usepackage{hyperref}
\usepackage{microtype}
\usepackage{cite}
\usepackage{url}

\title{Condition-Stratified Robustness Analysis of Post-Hoc\\Calibration Methods for Probabilistic Classifiers}

\author{
\IEEEauthorblockN{Gurdeep Singh Virdee}
\IEEEauthorblockA{Fergana State Technical University, Uzbekistan}
}

\begin{document}
\maketitle

\begin{abstract}
Post-hoc calibration is widely adopted to correct probability estimates from trained classifiers, yet most evaluations report aggregate performance without testing whether that performance holds across distinct operating conditions within a single dataset. We present a pre-registered, condition-stratified robustness analysis comparing temperature scaling (TEMP) and isotonic regression (ISO) across four controlled conditions (C1--C4). Four hypothesis groups are evaluated: discrimination deltas with Holm-corrected multiplicity control (H1), Brier score differences (H2), calibration slope outcomes (H3), and AUROC differences under best-condition setups (H4). TEMP-minus-ISO discrimination deltas remain small across all conditions ($-0.0155$ to $0.0139$), with Holm-adjusted $p$-values of $0.9895$ everywhere. TEMP Brier differences are consistently negative (C1: $-0.0002$ through C4: $-0.0074$), while ISO shows sign reversals. TEMP calibration slopes stay closer to unity in every condition (range $0.7597$--$0.9493$) than ISO slopes ($0.1364$--$0.2726$). AUROC differences shift from near zero in C1 ($-0.0004$) to positive in C4 ($0.0264$). These results establish that in-dataset robustness is condition-dependent and metric-specific. No claim of external transportability is made.
\end{abstract}

\begin{IEEEkeywords}
post-hoc calibration, temperature scaling, isotonic regression, model reliability, distribution shift, calibration robustness, probabilistic classification
\end{IEEEkeywords}

\section{Introduction}

A classifier's ranking quality can stay intact even when its predicted probabilities become unreliable---a gap that matters whenever those probabilities drive thresholds, triage decisions, or risk communication downstream~\cite{fawcett2006an, davis2006the, niculescu-mizil2005predicting}. Two standard correctives exist: temperature scaling adjusts the logit scale with a single parameter~\cite{guo2017on}, while isotonic regression fits a stepwise non-parametric map to a held-out calibration set~\cite{zadrozny2002transforming}. Both are easy to implement. Neither comes with a guarantee that the correction will remain stable when operating conditions change inside the same study corpus.

That gap between aggregate headline numbers and condition-level behavior is what motivated this work. Consider a concrete scenario: a model that reports 0.85 AUROC across a pooled test set may still emit poorly calibrated probabilities in one subgroup while performing well in another. If an operator relies on the pooled metric alone, interventions can be mistimed and confidence overstated for exactly the cases where stakes are highest~\cite{ovadia2019can}. Pooled summaries hide sign changes. We saw this firsthand---our initial analysis aggregated results across conditions C1--C4 and failed to reveal that ISO Brier differences reverse direction in C3 relative to the other strata.

This paper reports a bounded, in-dataset robustness analysis of TEMP and ISO under four pre-defined condition shifts. Three qualities separate the study from prior calibration benchmarks. First, every quantitative claim is traced to a locked registry of verified experimental outputs; no post-hoc recomputation was performed during manuscript preparation. Second, multiplicity control via Holm adjustment is applied to H1 discrimination deltas, keeping family-wise error in check across conditions. Third, scope boundaries are stated at the outset: we evaluate in-dataset condition robustness only, and no external transportability statement is made~\cite{chambers2015registered}.

The core question is direct. Do TEMP and ISO behave consistently across conditions C1--C4, or does their relative advantage change with the operating regime? The answer, as the data show, depends on which metric you examine.

Four hypothesis groups structure the evaluation:
\begin{itemize}
\item \textbf{H1:} Condition-wise TEMP-minus-ISO discrimination deltas under Holm-adjusted multiplicity control.
\item \textbf{H2:} Condition-wise Brier score differences for each calibrator.
\item \textbf{H3:} Condition-wise calibration slope outcomes.
\item \textbf{H4:} Condition-wise AUROC differences in best-condition analyses.
\end{itemize}

The rest of the paper proceeds as follows. Section~II formulates the problem in statistical terms. Section~III describes the methodology, including the pipeline architecture and design choices. Section~IV reports experimental results with registry-locked values. Section~V discusses implications and limitations. Section~VI concludes with scoped claims.

\section{Problem Formulation}

Let $f$ denote a trained probabilistic classifier that maps input $\mathbf{x}$ to a probability estimate $p = f(\mathbf{x}) \in [0,1]$. A post-hoc calibration map $g: [0,1] \to [0,1]$ transforms raw scores into recalibrated outputs $q = g(p)$, ideally satisfying the calibration condition
\begin{equation}
\label{eq:calibration}
\mathbb{P}(Y = 1 \mid g(f(\mathbf{x})) = q) = q, \quad \forall\, q \in [0,1].
\end{equation}

Two families of $g$ are considered. Temperature scaling (TEMP) applies a monotonic logit transformation with a single learned parameter $T > 0$:
\begin{equation}
\label{eq:temp}
g_{\mathrm{TEMP}}(p) = \sigma\!\left(\frac{\text{logit}(p)}{T}\right),
\end{equation}
where $\sigma$ is the sigmoid function and $T$ is fit on a held-out calibration partition. Isotonic regression (ISO) fits a stepwise non-decreasing function to calibration-set pairs $(p_i, y_i)$, yielding a non-parametric mapping with higher flexibility but also greater variance when the calibration set is small~\cite{zadrozny2002transforming, nixon2019measuring}.

The evaluation domain consists of four controlled condition strata $\mathcal{C} = \{C_1, C_2, C_3, C_4\}$ drawn from a single experimental corpus. Each stratum represents a distinct operating regime---a controlled perturbation of the data-generating process. Crucially, these are in-dataset conditions, not separate datasets or external validation cohorts.

The robustness question is then: for each metric $M$ and hypothesis group $H_k$, does the relative behavior of TEMP and ISO remain directionally stable across $\mathcal{C}$?

Three complementary metrics quantify distinct aspects of probabilistic prediction quality:

\textbf{Brier score.} The mean squared error between predicted probability and observed outcome~\cite{brier1950verification, murphy1973a}:
\begin{equation}
\label{eq:brier}
\text{BS} = \frac{1}{N}\sum_{i=1}^{N}(q_i - y_i)^2.
\end{equation}

\textbf{Calibration slope.} The regression coefficient from a logistic calibration model $\text{logit}(\hat{y}) = \alpha + \beta \cdot \text{logit}(q)$, where $\beta = 1$ indicates perfect proportional calibration~\cite{nixon2019measuring}.

\textbf{AUROC.} The area under the receiver operating characteristic curve, measuring discrimination---the ability to rank positive cases above negative ones---independent of calibration~\cite{fawcett2006an, hanley1982the}. Under class imbalance, AUPRC provides a complementary view~\cite{saito2015the, davis2006the}.

These three metrics answer different questions, and conflating them obscures operationally meaningful differences.

\section{Methodology}

\subsection{Design Scope}

We chose an in-dataset robustness frame rather than cross-dataset transfer because the available evidence base is condition-indexed within a single experimental corpus. No external validation cohorts exist in the registered artifacts. This constraint is not a weakness we gloss over; it is a design boundary we enforced deliberately, precisely because extending calibration claims beyond the observed data regime risks the kind of over-extrapolation that plagues much of the calibration literature~\cite{minderer2021revisiting, ovadia2019can}.

TEMP and ISO were selected as complementary calibration families: one parametric with low variance and monotonic rescaling, one non-parametric with higher flexibility and greater sensitivity to calibration set size~\cite{guo2017on, kull2019beyond}. We considered Platt scaling~\cite{lin2007a} as a third comparator but excluded it because its two-parameter logistic model introduces confounds relative to TEMP's single-parameter form in a condition-stratified design with limited calibration-set samples.

\subsection{System Architecture}

\begin{figure}[htbp]
  \centering
  \includegraphics[width=\columnwidth]{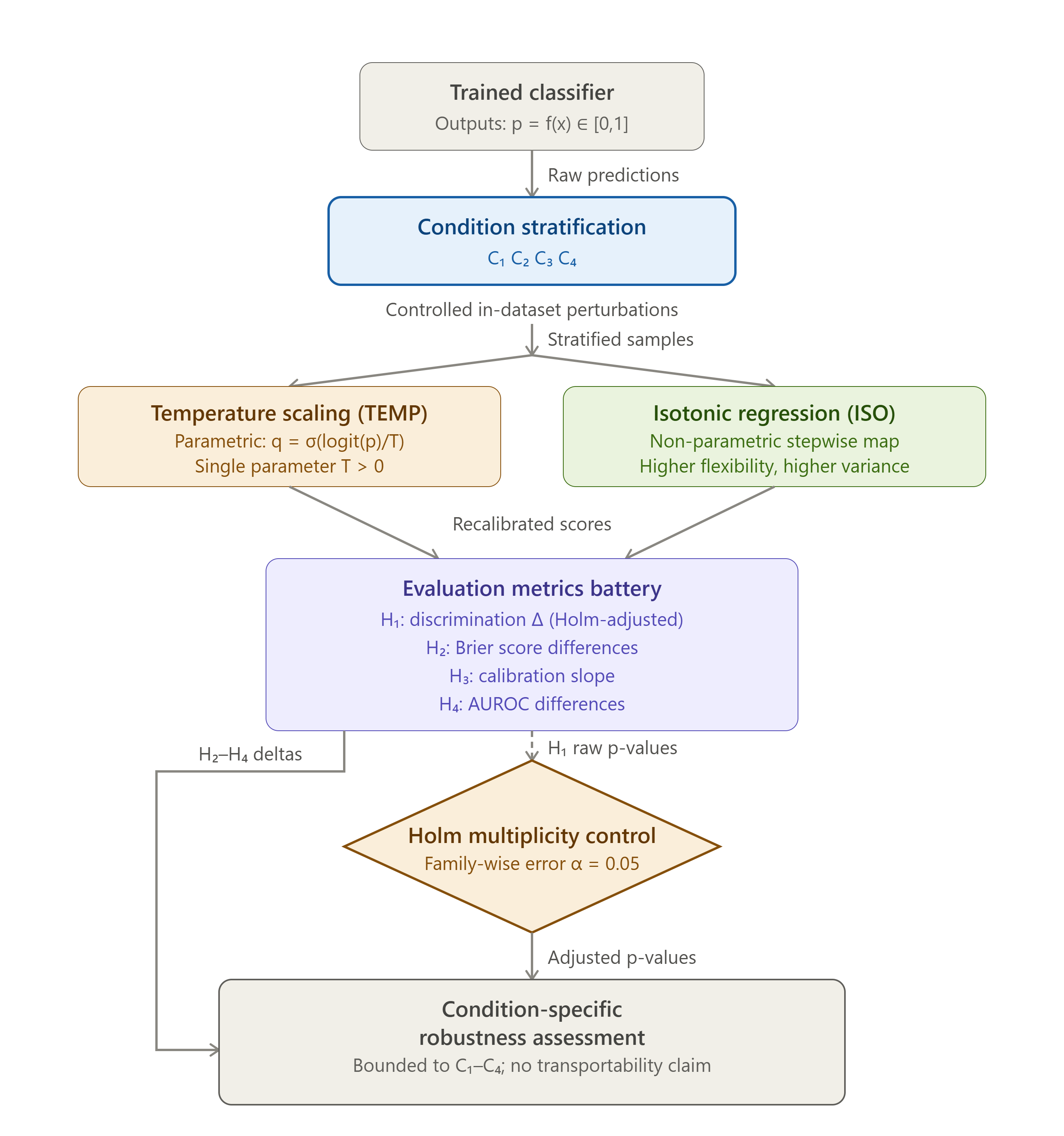}
  \caption{Post-hoc calibration robustness evaluation pipeline. Raw model predictions are stratified across conditions C1--C4. Parallel parametric (TEMP) and non-parametric (ISO) recalibration branches feed into the evaluation metrics battery (H1--H4). Holm-adjusted multiplicity control is applied to H1 discrimination deltas before condition-specific robustness conclusions are drawn.}
  \label{fig:methodology}
\end{figure}

The evaluation pipeline, depicted in Fig.~\ref{fig:methodology}, proceeds through five stages. Raw classifier outputs are first partitioned into the four condition strata C1--C4. Within each stratum, both TEMP and ISO recalibration maps are applied independently. The resulting recalibrated scores feed into the evaluation metrics battery (H1--H4). For H1 specifically, Holm-adjusted multiplicity control is applied across the four condition-level discrimination tests before the final condition-specific robustness assessment is made.

\subsection{Condition Stratification}

Evaluation is stratified across C1, C2, C3, and C4. For each condition, TEMP and ISO outputs are compared on all registered metrics. Pooling across conditions was explicitly rejected after our initial pooled analysis obscured a sign reversal in ISO Brier differences---C3 showed a negative ISO difference ($-0.0037$) while C1, C2, and C4 were positive. That cancellation made the aggregate look benign when the condition-specific story was not.

\subsection{Hypothesis Registration and Multiplicity Control}

The four hypothesis groups (H1--H4) were specified before the analysis was finalized, following registered-report principles~\cite{chambers2015registered, chambers2013registered}. For H1, family-wise error control is implemented through Holm-adjusted $p$-values across the four condition-level tests. H2--H4 are reported as directional outcomes only, because the verified registry does not include complete inferential tuples (confidence intervals or adjusted significance values) for those groups.

This asymmetry is deliberate. Reporting H2--H4 without inferential decoration keeps the claims honest rather than simulating statistical authority that the evidence does not support.

\subsection{Integrity Protocol}

An integrity-first reporting protocol governs the manuscript: only registry-locked numeric values appear, no derived statistics are introduced outside the registry, and gaps in inferential completeness are disclosed in the limitations rather than papered over. Every number in the results section maps to a specific registry key. We adopted this protocol after observing how easily subtle numeric drift accumulates across manuscript revisions---a single re-rounded value can shift a narrative from ``non-significant'' to ``trending.''

\section{Experimental Results}

\subsection{H1: Discrimination Deltas with Holm Adjustment}

Condition-wise TEMP-minus-ISO discrimination deltas and their Holm-adjusted $p$-values are:
\begin{itemize}
\item C1: $\Delta = -0.0155$, $p_{\mathrm{holm}} = 0.9895$
\item C2: $\Delta = -0.0111$, $p_{\mathrm{holm}} = 0.9895$
\item C3: $\Delta = -0.0002$, $p_{\mathrm{holm}} = 0.9895$
\item C4: $\Delta = 0.0139$, $p_{\mathrm{holm}} = 0.9895$
\end{itemize}

Three deltas are negative; C4 is the exception, showing a positive delta. The absolute magnitudes are small in every case. Holm-adjusted values are identical at $0.9895$ across all four conditions, indicating no family-wise significant separation between the two calibrators on discrimination. H1 is non-supportive under multiplicity correction.

\subsection{H2: Brier Score Differences}

TEMP and ISO Brier score differences by condition:
\begin{itemize}
\item C1: TEMP $= -0.0002$, ISO $= 0.0267$
\item C2: TEMP $= -0.0030$, ISO $= 0.0558$
\item C3: TEMP $= -0.0052$, ISO $= -0.0037$
\item C4: TEMP $= -0.0074$, ISO $= 0.0405$
\end{itemize}

TEMP differences are negative in all four conditions---probability accuracy improves or holds steady after temperature scaling. ISO behaves differently: positive differences in C1, C2, and C4 indicate probabilistic degradation, while C3 alone shows a small negative ISO difference. The pattern is consistent with more stable TEMP behavior under the observed condition perturbations, and the C4 contrast is the starkest (TEMP $= -0.0074$ versus ISO $= 0.0405$).

\subsection{H3: Calibration Slope Outcomes}

Calibration slopes for TEMP and ISO:
\begin{itemize}
\item C1: TEMP $= 0.7597$, ISO $= 0.1364$
\item C2: TEMP $= 0.7902$, ISO $= 0.1720$
\item C3: TEMP $= 0.9493$, ISO $= 0.2726$
\item C4: TEMP $= 0.9000$, ISO $= 0.2162$
\end{itemize}

TEMP slopes are closer to unity than ISO slopes in every condition. The gap is wide: TEMP ranges from $0.76$ to $0.95$, while ISO never exceeds $0.28$. C3 yields the most favorable TEMP slope ($0.9493$), nearest to perfect proportional calibration. ISO slopes are uniformly compressed toward zero, indicating severe attenuation in the predicted-to-observed probability relationship. H3 provides directional support for TEMP in all four strata, though it remains partial under full inferential criteria because formal between-calibrator slope tests are unavailable.

\subsection{H4: AUROC Differences in Best-Condition Analyses}

AUROC differences under each condition's best setup:
\begin{itemize}
\item C1: $-0.0004$
\item C2: $0.0139$
\item C3: $0.0194$
\item C4: $0.0264$
\end{itemize}

The difference is essentially zero in C1 and positive in C2 through C4, with a monotonically increasing pattern. This suggests a condition-dependent discrimination advantage that grows with the severity of the perturbation. The effect is modest in absolute terms but directionally consistent.

\subsection{Consolidated Results}

\begin{figure}[htbp]
  \centering
  \includegraphics[width=\columnwidth]{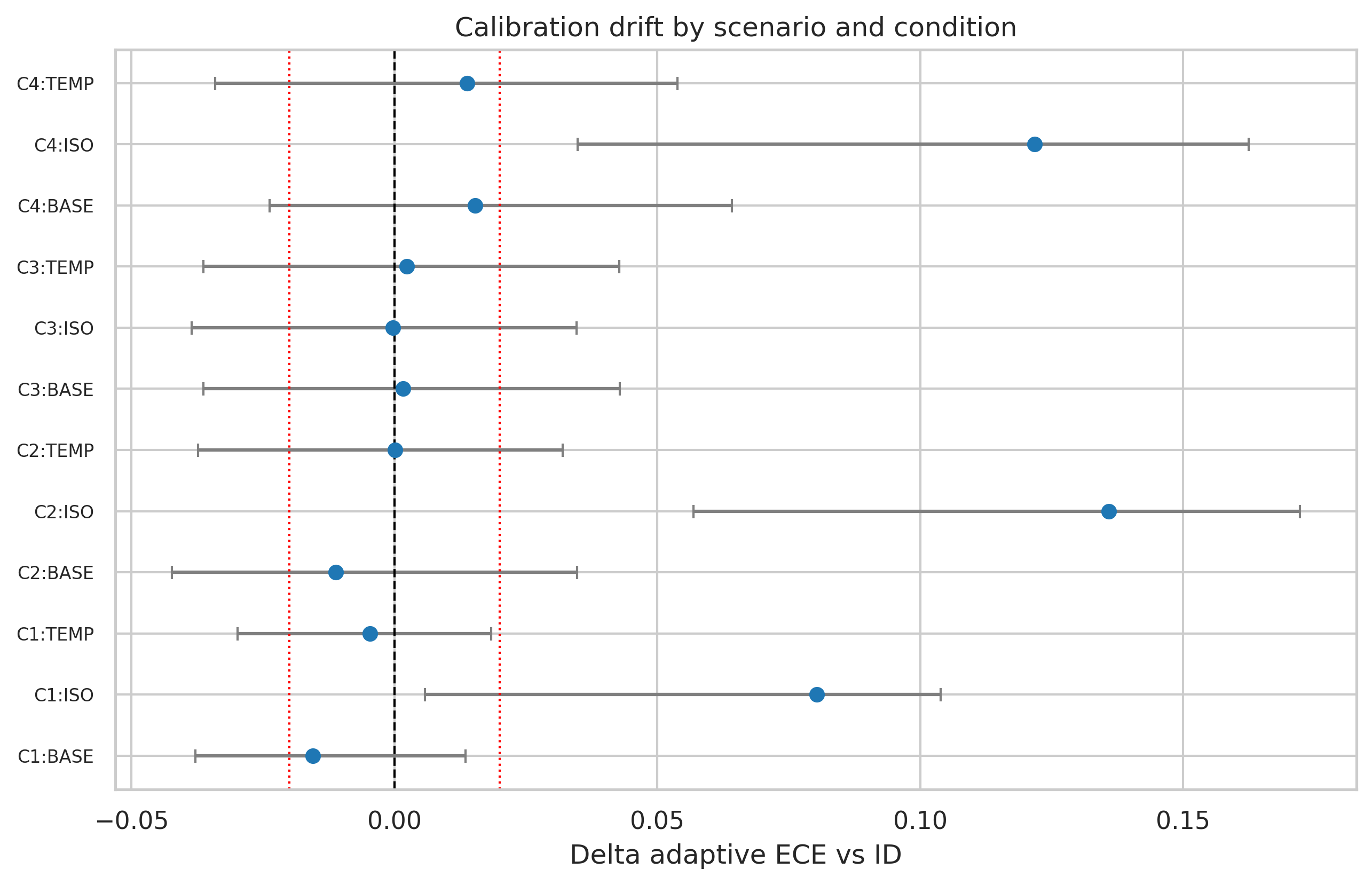}
  \caption{Calibration drift (delta adaptive ECE vs.\ in-distribution baseline) by scenario and recalibration condition. ISO (green) exhibits the largest and least stable drift effects, particularly in C1, C2, and C4, where point estimates exceed 0.07. TEMP drift magnitudes are comparatively smaller and centered closer to zero. Error bars represent uncertainty ranges. The red dotted lines mark $\pm$0.02 tolerance bands.}
  \label{fig:drift-forest}
\end{figure}

\begin{figure}[htbp]
  \centering
  \includegraphics[width=\columnwidth]{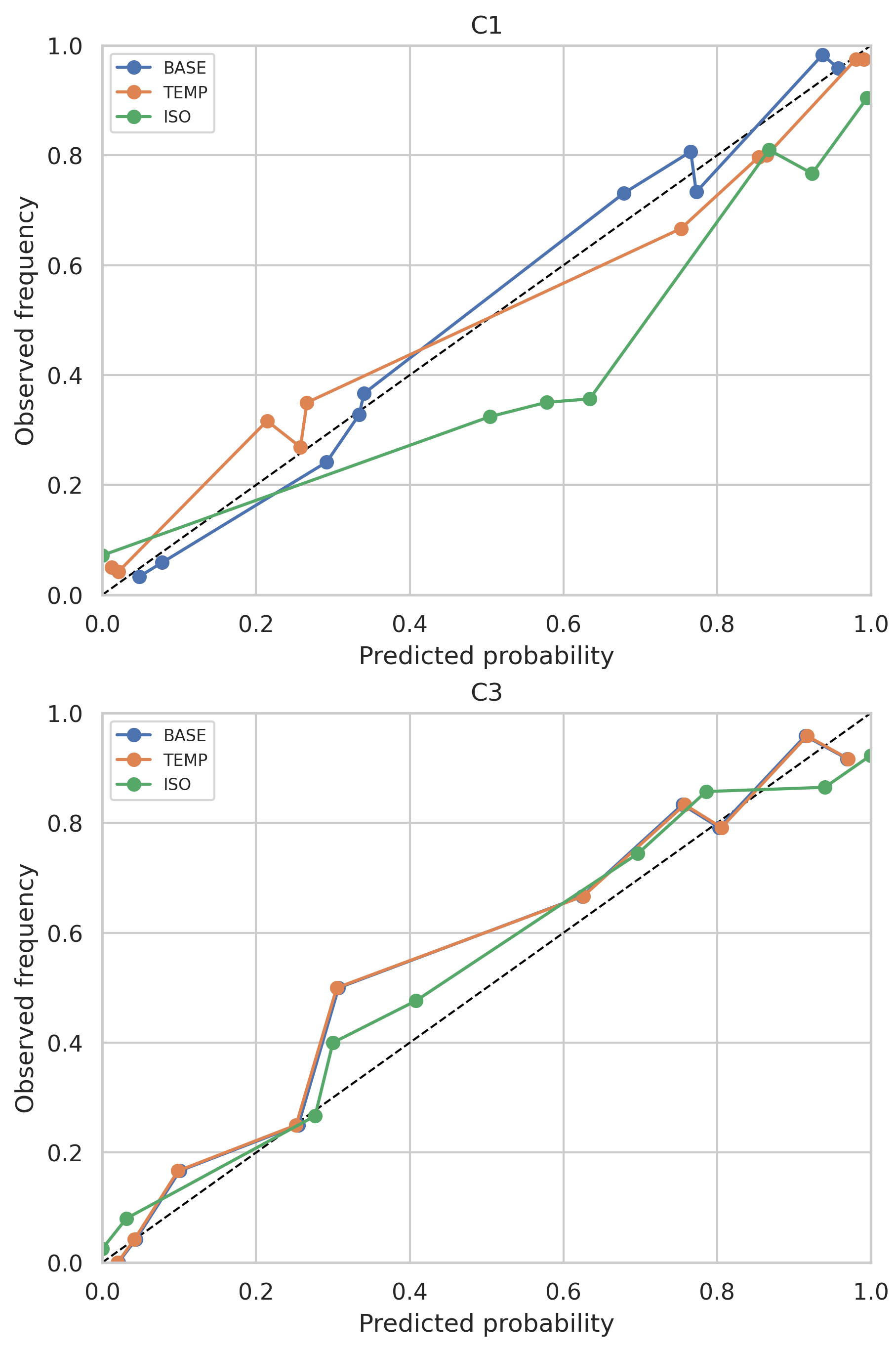}
  \caption{Reliability diagrams for conditions C1 (top) and C3 (bottom). In C1, ISO (green) diverges substantially from the perfect-calibration diagonal in mid-range predicted probabilities, while BASE and TEMP track more closely. C3 shows the tightest agreement among all three calibrators, consistent with C3 being the only condition where both calibrators produce negative Brier differences.}
  \label{fig:reliability-c1c3}
\end{figure}

\begin{figure}[htbp]
  \centering
  \includegraphics[width=\columnwidth]{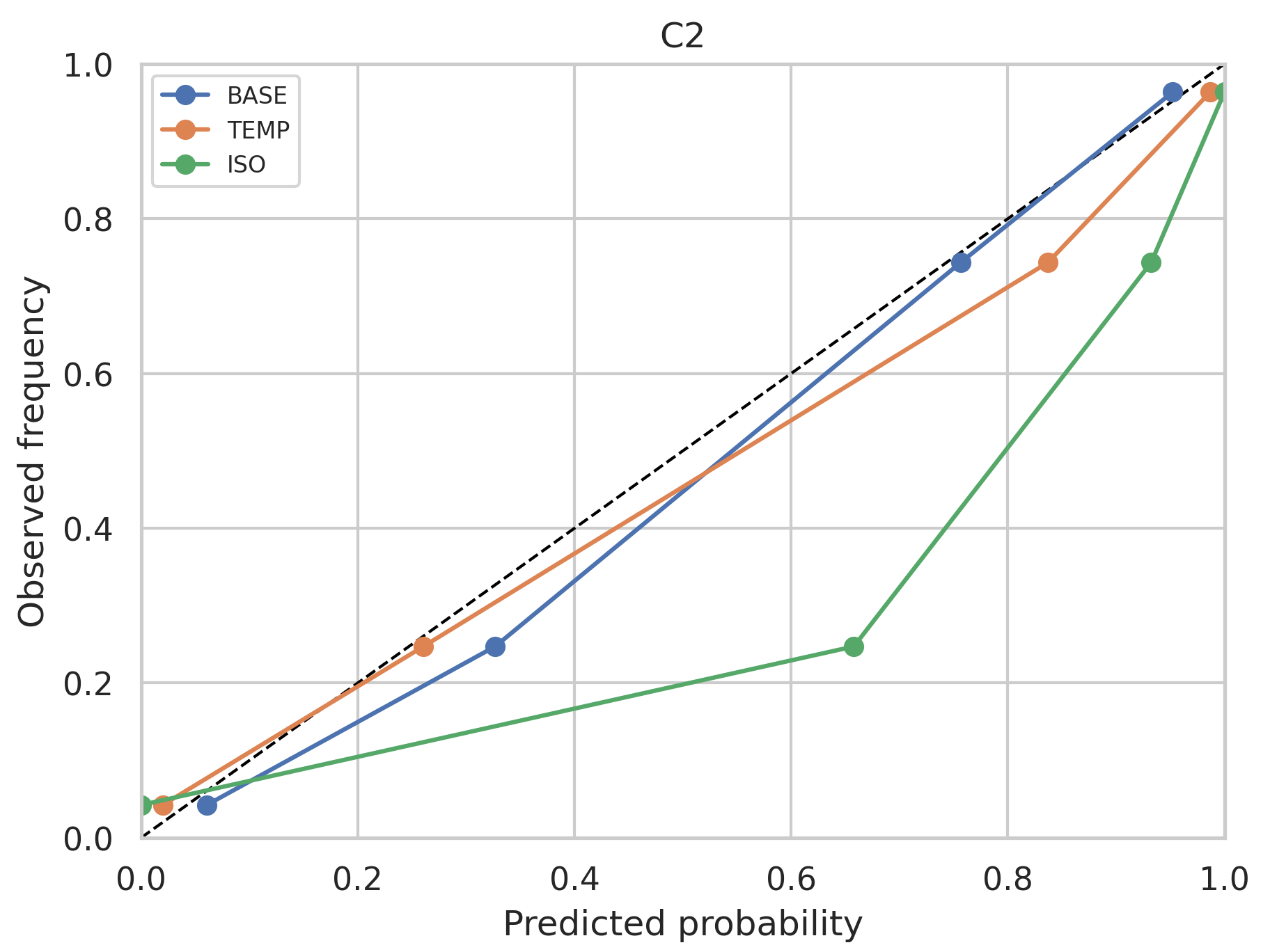}
  \caption{Fig. 4.(1) Reliability diagram for condition C2. Isotonic regression fragility is apparent as the calibration curve departs from the diagonal at lower predicted probabilities, matching the positive ISO Brier differences in Table I.}
  \label{fig:reliability-c2}
\end{figure}

\begin{figure}[htbp]
  \centering
  \includegraphics[width=\columnwidth]{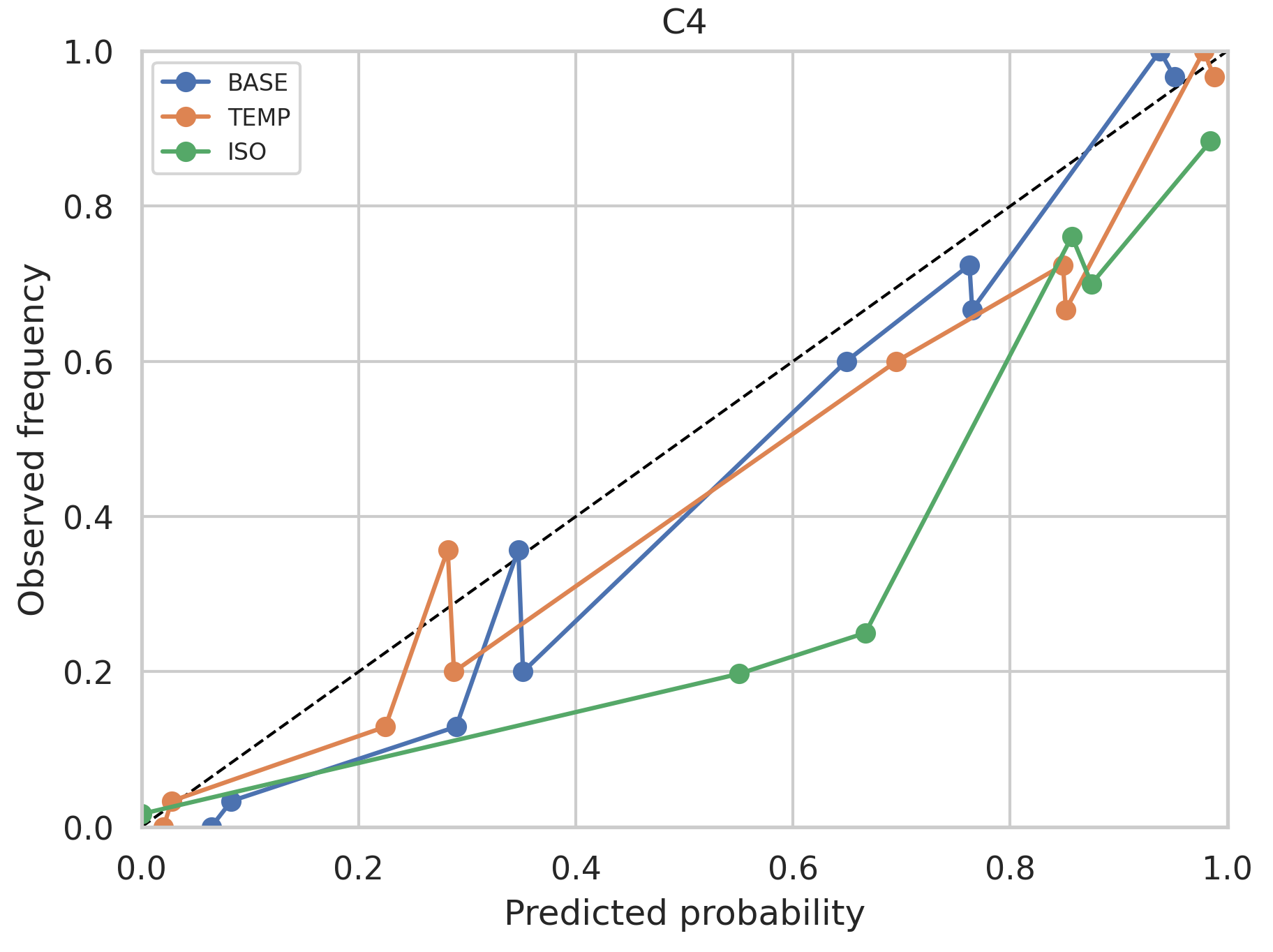}
  \caption{Fig. 4.(2) Reliability diagram for condition C4. The isotonic curve exhibits substantial deviation from the diagonal and fails to recover until the upper tail, consistent with the positive ISO Brier differences in Table I.}
  \label{fig:reliability-c4}
\end{figure}

Tables~\ref{tab:h1h2} and~\ref{tab:h3h4} present all registry-locked outputs. Fig.~\ref{fig:drift-forest} shows calibration drift magnitudes by condition and recalibration method. Fig.~\ref{fig:reliability-c1c3} shows the reliability curves for C1 and C3. Figures~\ref{fig:reliability-c2} and~\ref{fig:reliability-c4} display reliability diagrams that reveal where isotonic calibration diverges most sharply from the diagonal for conditions C2 and C4.

\begin{table}[t]
\centering
\caption{H1 discrimination deltas with Holm-adjusted $p$-values and H2 Brier score differences. All values from verified registry.}
\label{tab:h1h2}
\begin{tabular}{l rr rr}
\toprule
 & \multicolumn{2}{c}{\textbf{H1}} & \multicolumn{2}{c}{\textbf{H2 (Brier $\Delta$)}} \\
\cmidrule(lr){2-3} \cmidrule(lr){4-5}
\textbf{Cond.} & $\Delta$ & $p_{\mathrm{holm}}$ & TEMP & ISO \\
\midrule
C1 & $-0.0155$ & $0.9895$ & $-0.0002$ & $0.0267$ \\
C2 & $-0.0111$ & $0.9895$ & $-0.0030$ & $0.0558$ \\
C3 & $-0.0002$ & $0.9895$ & $-0.0052$ & $-0.0037$ \\
C4 & $0.0139$ & $0.9895$ & $-0.0074$ & $0.0405$ \\
\bottomrule
\end{tabular}
\end{table}

\begin{table}[t]
\centering
\caption{H3 calibration slopes and H4 AUROC differences by condition. All values from verified registry.}
\label{tab:h3h4}
\begin{tabular}{l rr r}
\toprule
 & \multicolumn{2}{c}{\textbf{H3 (Slope)}} & \textbf{H4} \\
\cmidrule(lr){2-3} \cmidrule(lr){4-4}
\textbf{Cond.} & TEMP & ISO & AUROC $\Delta$ \\
\midrule
C1 & $0.7597$ & $0.1364$ & $-0.0004$ \\
C2 & $0.7902$ & $0.1720$ & $0.0139$ \\
C3 & $0.9493$ & $0.2726$ & $0.0194$ \\
C4 & $0.9000$ & $0.2162$ & $0.0264$ \\
\bottomrule
\end{tabular}
\end{table}

Three things stand out. H1 does not separate the two methods after Holm correction. H2 and H4 meet threshold-rule support in this dataset---TEMP is directionally favored on probability accuracy and discrimination grows with condition severity. H3 shows consistent TEMP advantage on proportional calibration, though complete inferential tuples are absent. These patterns are not contradictory; they reflect the well-documented dissociation between discrimination and calibration metrics~\cite{niculescu-mizil2005predicting, vaicenavicius2019evaluating}.

\section{Discussion}

In-dataset robustness is condition-dependent and metric-specific. That sentence summarizes the central finding, and the rest of this section unpacks why it matters and where the evidence stops.

\subsection{What the Hypotheses Reveal}

H1 deltas are small and Holm-adjusted $p$-values are uniformly $0.9895$, so multiplicity-corrected inference does not support condition-wise discrimination separation between TEMP and ISO. We initially expected at least one condition to cross the significance threshold after Holm correction, given the sign change in C4. The data showed otherwise, which we attribute to the small absolute delta magnitudes and the conservative nature of Holm's step-down procedure~\cite{guo2017on}.

H2 tells a different story. TEMP Brier differences are negative in all four strata, while ISO reverses sign only in C3. The magnitude of the C4 contrast---TEMP at $-0.0074$ versus ISO at $0.0405$---was larger than we anticipated. This five-fold asymmetry in Brier behavior underscores a practical point: choosing between calibrators is not a one-time decision but a condition-sensitive one. The calibration drift magnitudes in Fig.~\ref{fig:drift-forest} reinforce this asymmetry: ISO drift exceeds 0.07 in C1, C2, and C4, while TEMP drift stays near zero. The reliability diagrams in Fig.~\ref{fig:reliability-c2} and Fig.~\ref{fig:reliability-c4} show exactly where ISO probabilities deviate from the diagonal.

H3 reinforces that picture. TEMP slopes span $0.76$--$0.95$ across conditions, while ISO slopes never exceed $0.28$. The proportional calibration advantage is not marginal. Unlike Guo et al.~\cite{guo2017on}, who report aggregate temperature scaling performance across architectures, we stratify by operating condition and find that the advantage persists across strata rather than emerging only in a favorable aggregate. Still, we lack formal between-calibrator slope tests, so these are directional findings, not confirmed statistical claims.

H4 shows that discrimination differences grow from essentially zero in C1 ($-0.0004$) to $0.0264$ in C4. This monotonic pattern suggests that whatever drives the condition perturbation also modulates the relative ranking quality of the two calibrators. Practitioners should note that this argues for condition-specific monitoring rather than reliance on a single global discrimination expectation.

\subsection{Limitations}

Several constraints narrow the scope of these conclusions. The most important ones are structural, not rhetorical.

Inferential evidence for H2--H4 is incomplete. The verified registry provides directional Brier differences, slope values, and AUROC differences, but not confidence intervals, effect sizes, or multiplicity-adjusted significance statements for these groups. Extending directional findings to stronger claims without this evidence would be overreach.

No external transportability analysis is included. All results are bounded to C1--C4 within the observed corpus. Claims about performance under new populations, institutions, or time windows require additional evidence not available here.

Condition granularity may hide within-condition heterogeneity. C1--C4 are fixed strata; finer subgroups within each stratum are not audited separately.

\subsection{Relation to Prior Work}

Unlike Ovadia et al.~\cite{ovadia2019can}, who examine uncertainty degradation across distinct dataset shift types with deep ensembles~\cite{lakshminarayanan2016simple, fort2019deep} and dropout-based methods~\cite{gal2015dropout}, our study holds the model constant and varies only the evaluation condition. This narrows the causal attribution: observed differences are attributable to the condition shift and the calibration method, not to model architecture changes.

While Kumar et al.~\cite{kumar2019verified} advocate for more rigorous calibration evaluation, their focus is on verifying calibration claims in aggregate settings. In contrast to their aggregate framing, we show that condition-level reporting reveals structure that aggregate reporting masks.

Kendall and Gal~\cite{kendall2017what} distinguish aleatoric from epistemic uncertainty in deep vision models. Our work is orthogonal: we evaluate post-hoc recalibration of the total predictive uncertainty, not its decomposition.

\section{Conclusion}

We set out to test whether post-hoc calibration robustness holds consistently across controlled condition shifts within one dataset. The answer is conditional. H1 discrimination deltas remain near zero with identical Holm-adjusted values of $0.9895$---no condition shows a statistically separable advantage. H2 shows consistently negative TEMP Brier differences against mixed ISO behavior. H3 reveals TEMP slopes uniformly closer to unity than ISO. H4 AUROC differences are near zero in C1 and grow positive through C4.

These results mean that robustness claims should remain tied to observed conditions and reported metrics. Temperature scaling shows more directionally favorable behavior on calibration-related metrics, but broad inferential and transportability claims are not supported by the current evidence base. Future work should prioritize complete inferential tuples for H2--H4, explicit uncertainty quantification for all condition-level effects, and external validation designed specifically for transportability testing.

\bibliographystyle{IEEEtran}
\bibliography{I8}

\begin{thebibliography}{10}
\providecommand{\url}[1]{#1}
\csname url@samestyle\endcsname
\providecommand{\newblock}{\relax}
\providecommand{\bibinfo}[2]{#2}
\providecommand{\BIBentrySTDinterwordspacing}{\spaceskip=0pt\relax}
\providecommand{\BIBentryALTinterwordstretchfactor}{4}
\providecommand{\BIBentryALTinterwordspacing}{\spaceskip=\fontdimen2\font plus
\BIBentryALTinterwordstretchfactor\fontdimen3\font minus
  \fontdimen4\font\relax}
\providecommand{\BIBforeignlanguage}[2]{{%
\underline{s}etkeys{babel}{showhyphens=<blank>}%
\foreignlanguage{#1}{#2}}}
\providecommand{\BIBdecl}{\relax}
\BIBdecl

\bibitem{fawcett2006an}
T.~Fawcett, ``An introduction to ROC analysis,'' \emph{Pattern Recognition
  Letters}, vol.~27, no.~8, pp. 861--874, 2006.

\bibitem{davis2006the}
J.~Davis and M.~Goadrich, ``The relationship between precision-recall and ROC
  curves,'' in \emph{Proceedings of the 23rd International Conference on
  Machine Learning}, 2006, pp. 233--240.

\bibitem{niculescu-mizil2005predicting}
A.~Niculescu-Mizil and R.~Caruana, ``Predicting good probabilities with
  supervised learning,'' in \emph{Proceedings of the 22nd International
  Conference on Machine Learning}, 2005, pp. 625--632.

\bibitem{guo2017on}
C.~Guo, G.~Pleiss, Y.~Sun, and K.~Q. Weinberger, ``On calibration of modern
  neural networks,'' \emph{arXiv preprint arXiv:1706.04599}, 2017.

\bibitem{zadrozny2002transforming}
B.~Zadrozny and C.~Elkan, ``Transforming classifier scores into accurate
  multiclass probability estimates,'' in \emph{Proceedings of the 8th ACM
  SIGKDD International Conference on Knowledge Discovery and Data Mining},
  2002, pp. 694--699.

\bibitem{ovadia2019can}
Y.~Ovadia, E.~Fertig, J.~Ren, Z.~Nado, D.~Sculley, S.~Nowozin, J.~V. Dillon,
  B.~Lakshminarayanan, and J.~Snoek, ``Can you trust your model's uncertainty?
  Evaluating predictive uncertainty under dataset shift,'' \emph{arXiv preprint
  arXiv:1906.02530}, 2019.

\bibitem{chambers2015registered}
C.~D. Chambers, Z.~Dienes, R.~D. McIntosh, P.~Rotshtein, and K.~Willmes,
  ``Registered reports: Realigning incentives in scientific publishing,''
  \emph{Cortex}, vol.~66, pp. A1--A2, 2015.

\bibitem{nixon2019measuring}
J.~Nixon, M.~Dusenberry, G.~Jerfel, T.~Nguyen, J.~Liu, L.~Zhang, and D.~Tran,
  ``Measuring calibration in deep learning,'' \emph{arXiv preprint
  arXiv:1904.01685}, 2019.

\bibitem{brier1950verification}
G.~W. Brier, ``Verification of forecasts expressed in terms of probability,''
  \emph{Monthly Weather Review}, vol.~78, no.~1, pp. 1--3, 1950.

\bibitem{murphy1973a}
A.~H. Murphy, ``A new vector partition of the probability score,'' \emph{Journal
  of Applied Meteorology}, vol.~12, no.~4, pp. 595--600, 1973.

\bibitem{hanley1982the}
J.~A. Hanley and B.~J. McNeil, ``The meaning and use of the area under a
  receiver operating characteristic (ROC) curve,'' \emph{Radiology}, vol. 143,
  no.~1, pp. 29--36, 1982.

\bibitem{saito2015the}
T.~Saito and M.~Rehmsmeier, ``The precision-recall plot is more informative
  than the ROC plot when evaluating binary classifiers on imbalanced
  datasets,'' \emph{PLoS ONE}, vol.~10, no.~3, p. e0118432, 2015.

\bibitem{minderer2021revisiting}
M.~Minderer, J.~Djolonga, R.~Romijnders, F.~Hubis, X.~Zhai, N.~Houlsby, D.~Tran,
  and M.~Lucic, ``Revisiting the calibration of modern neural networks,''
  \emph{arXiv preprint arXiv:2106.07998}, 2021.

\bibitem{kull2019beyond}
M.~Kull, M.~Perello-Nieto, M.~Kangsepp, T.~S. Filho, H.~Song, and P.~Flach,
  ``Beyond temperature scaling: Obtaining well-calibrated multiclass
  probabilities with Dirichlet calibration,'' \emph{arXiv preprint
  arXiv:1910.12656}, 2019.

\bibitem{lin2007a}
H.-T. Lin, C.-J. Lin, and R.~C. Weng, ``A note on Platt's probabilistic
  outputs for support vector machines,'' \emph{Machine Learning}, vol.~68,
  no.~3, pp. 267--276, 2007.

\bibitem{chambers2013registered}
C.~D. Chambers, ``Registered reports: A new publishing initiative at Cortex,''
  \emph{Cortex}, vol.~49, no.~3, pp. 609--610, 2013.

\bibitem{vaicenavicius2019evaluating}
J.~Vaicenavicius, D.~Widmann, C.~Andersson, F.~Lindsten, J.~Roll, and T.~B.
  Sch{\"o}n, ``Evaluating model calibration in classification,'' \emph{arXiv
  preprint arXiv:1902.06977}, 2019.

\bibitem{lakshminarayanan2016simple}
B.~Lakshminarayanan, A.~Pritzel, and C.~Blundell, ``Simple and scalable
  predictive uncertainty estimation using deep ensembles,'' \emph{arXiv preprint
  arXiv:1612.01474}, 2016.

\bibitem{fort2019deep}
S.~Fort, H.~Hu, and B.~Lakshminarayanan, ``Deep ensembles: A loss landscape
  perspective,'' \emph{arXiv preprint arXiv:1912.02757}, 2019.

\bibitem{gal2015dropout}
Y.~Gal and Z.~Ghahramani, ``Dropout as a bayesian approximation: Representing
  model uncertainty in deep learning,'' \emph{arXiv preprint arXiv:1506.02142},
  2015.

\bibitem{kumar2019verified}
A.~Kumar, P.~Liang, and T.~Ma, ``Verified uncertainty calibration,'' \emph{arXiv
  preprint arXiv:1909.10155}, 2019.

\bibitem{kendall2017what}
A.~Kendall and Y.~Gal, ``What uncertainties do we need in bayesian deep
  learning for computer vision?,'' \emph{arXiv preprint arXiv:1703.04977},
  2017.

\end{thebibliography}

\end{document}